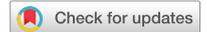

# VME: A Satellite Imagery Dataset and Benchmark for Detecting Vehicles in the Middle East and Beyond

Noora Al-Emadi[1,2 ✉], Ingmar Weber[3], Yin Yang[2] & Ferda Ofli[1]

Detecting vehicles in satellite images is crucial for traffic management, urban planning, and disaster response. However, current models struggle with real-world diversity, particularly across different regions. This challenge is amplified by geographic bias in existing datasets, which often focus on specific areas and overlook regions like the Middle East. To address this gap, we present the Vehicles in the Middle East (VME) dataset, designed explicitly for vehicle detection in high-resolution satellite images from Middle Eastern countries. Sourced from Maxar, the VME dataset spans 54 cities across 12 countries, comprising over 4,000 image tiles and more than 100,000 vehicles, annotated using both manual and semi-automated methods. Additionally, we introduce the largest benchmark dataset for Car Detection in Satellite Imagery (CDSI), combining images from multiple sources to enhance global car detection. Our experiments demonstrate that models trained on existing datasets perform poorly on Middle Eastern images, while the VME dataset significantly improves detection accuracy in this region. Moreover, state-of-the-art models trained on CDSI achieve substantial improvements in global car detection.

## Background & Summary

Satellite imagery has become an essential instrument for a wide range of applications from agriculture[1] and environmental monitoring[2] to urban development[3,4] and disaster response[5]. A recent review of object detection in satellite imagery highlights the difficulty of creating a general-purpose model that can handle thousands of diverse object categories and varying real-world conditions[6]. Instead, the study recommends focusing on task-specific models in narrower application areas, where success is more likely if large, well-annotated datasets are available. Therefore, our study focuses on vehicle detection in satellite imagery, a critical task with diverse real-world applications such as analyzing traffic flow and patterns for traffic management[7,8], monitoring parking lot occupancy rates to support urban planning[9], and modeling spatial-temporal changes in vehicle counts as a proxy for internal displacement monitoring[10]. To this end, we first present a novel labeled dataset called Vehicles in the Middle East (VME) to attenuate the under-representation of the region. We then construct the largest benchmark dataset, called Car Detection in Satellite Imagery (CDSI), by consolidating images from multiple existing satellite imagery datasets for enhanced global car detection.

Detecting vehicles in satellite imagery is challenging because each vehicle covers only a few pixels, classifying them as tiny objects. As a result, the surrounding context becomes crucial for accurately delineating these small objects. Several studies have been conducted on tiny object detection in satellite imagery[11–13]. A review comprehensively analyzed these methods based on five factors: data augmentation, multi-scale feature learning, context-based detection, training strategy, and GAN-based detection, and showed that these factors play a role in enhancing the detection performance in tiny objects[14]. Another systematic study on small object detection was conducted by reviewing existing literature on algorithms and datasets[15]. Two large-scale benchmarks, SODA-D and SODA-A, were constructed for driving scenarios and aerial scenes. Several algorithms were evaluated on top of these benchmarks with in-depth analyses, resulting in discussions on backbone effectiveness, hierarchical feature representation efficiency, and one-stage detector performance for small object detection. In addition, several studies were performed on vehicle and car detection[16–19]. These studies[17,18] focus on the development of new vehicle detection models, as well as the enhancement of existing ones, utilizing publicly available datasets such as DOTA[20], VEDAI[21], and xView[22], fMoW[23], VAID[24], and AI-TOD[25].

[1]Qatar Computing Research Institute, Hamad Bin Khalifa University, Doha, Qatar. [2]College of Science and Engineering, Hamad Bin Khalifa University, Doha, Qatar. [3]Saarland Informatics Campus, Saarland University, Saarbrücken, Germany. ✉e-mail: nalemadi@hbku.edu.qa

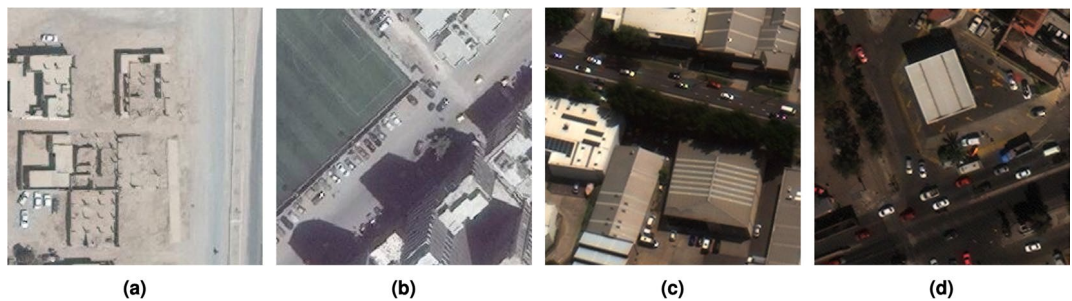

**Fig. 1** The distinct visual context of cars on the road in the Middle Eastern cities: (**a**) Abu Kamal, Syria, and (**b**) Alexandria, Egypt & other cities around the world: (**c**) Sydney, Australia[22], and (**d**) Mexico City, Mexico[22].

However, existing models for vehicle detection face challenges when applied to diverse real-world scenarios involving the analysis of satellite images from previously unexplored geographic regions[26]. For example, the visual context of a car on the road in Abu Kamal City, Syria (Fig. 1a) and Alexandria City, Egypt (Fig. 1b) present clear differences compared to a car on the road in Sydney, Australia (Fig. 1c) and Mexico City, Mexico (Fig. 1d). A noticeable contrast is evident in the appearance of built structures and land cover, stemming from unique differences in the natural landscape, climate, economic development, urban planning, and architectural design in Middle Eastern countries. This contrast becomes more pronounced thanks to the rapid pace of urban development in Middle Eastern countries driven by large-scale smart city projects as opposed to the more incremental urban upgrades seen in the US and Europe[27,28].

Therefore, with the prevalence of datasets focusing on specific regions, a gap related to geographic bias has emerged, particularly in the Middle East as highlighted in Fig. 2. To bridge this gap, the VME dataset, collected from Maxar, spans 54 cities in 12 countries in the Middle East and comprises more than 4,000 high-resolution image tiles of 512 × 512 pixels with more than 100k vehicle instances. The ground-truth annotations were generated using a combination of manual annotation and semi-automated techniques through a crowdsourcing company. Additionally, the CDSI dataset constitutes the largest benchmark for car detection by expanding VME with images from other existing satellite imagery datasets, such as xView[22], DOTA-v2.0[20], VEDAI[21], DIOR[29], and FAIR1M-2.0[30].

We conduct comprehensive experiments using advanced object detection models, such as TOOD[31] and DINO[32], and present baseline results on both individual and combined datasets. The VME baseline evaluation demonstrates a remarkable 56.3% improvement in mAP for car detection in the Middle East compared to models trained on existing datasets. Additionally, the model trained on the CDSI dataset, due to its greater diversity and scale, significantly enhances mAP50, with improvements ranging from 19.6% to 84.6% across all models trained on individual datasets. This newly developed dataset serves as a valuable resource for researchers and professionals in remote sensing, promoting progress in vehicle detection and satellite imagery analysis.

## Methods

This section provides details about our novel VME dataset such as the different categories, image resolution, area coverage, and annotation format. Then, we elaborate on the new benchmark dataset (CDSI), where we collect car-related objects from the publicly available datasets and combine them with the VME dataset.

**VME Dataset.** We constructed the VME dataset by collecting satellite images of different cities in the Middle Eastern countries such as Syria, Libya, Iraq, Jordan, Egypt, Qatar, Saudi Arabia, United Arab Emirates, Oman, Kuwait, and Bahrain. We included the most popular cities including the capitals of these countries. The city-level geographic distribution of the collected images in the VME dataset is highlighted with purple circles in Fig. 2, which includes underrepresented geographic regions for vehicle detection in satellite imagery, compared to the blue circles representing the distribution of images in the xView dataset. We note that the remaining datasets do not provide any geographical information at the country or city level and, hence, cannot be accurately represented on the map.

*Image Collection.* For each city in our dataset, we identified the geographic area of interest (AOI) and collected high-resolution satellite images from Maxar Technologies, which provides access to a large archive of the world's most recent pan-sharpened natural color images at a spatial resolution of up to 30 cm through a paid subscription to their SecureWatch platform. To this end, we searched the archive for satellite images with (i) RGB color, (ii) less than 20% cloud coverage, (iii) a ground sampling distance of at most 50 cm (i.e., images at 30 cm, 40 cm, and 50 cm spatial resolution), and (iv) off-Nadir angle less than 30 degrees.

We downloaded a total of 2,714 image snapshots across all 54 city AOIs. The resulting images are large with an average dimension in the range of 22,475 × 24,043. Since this image size is too large for processing and labeling directly, we generated random crops of image tiles with 512 × 512 pixels. This initially yielded a total of 22,125 image tiles. We ensured the resulting tiles did not have any missing or undefined pixels. Furthermore, to keep the annotation budget under control, we manually discarded the tiles that did not have obvious objects to annotate, such as images with completely green or desert areas. As a result of this filtering, we had 4,303 image tiles to be annotated in the next step.

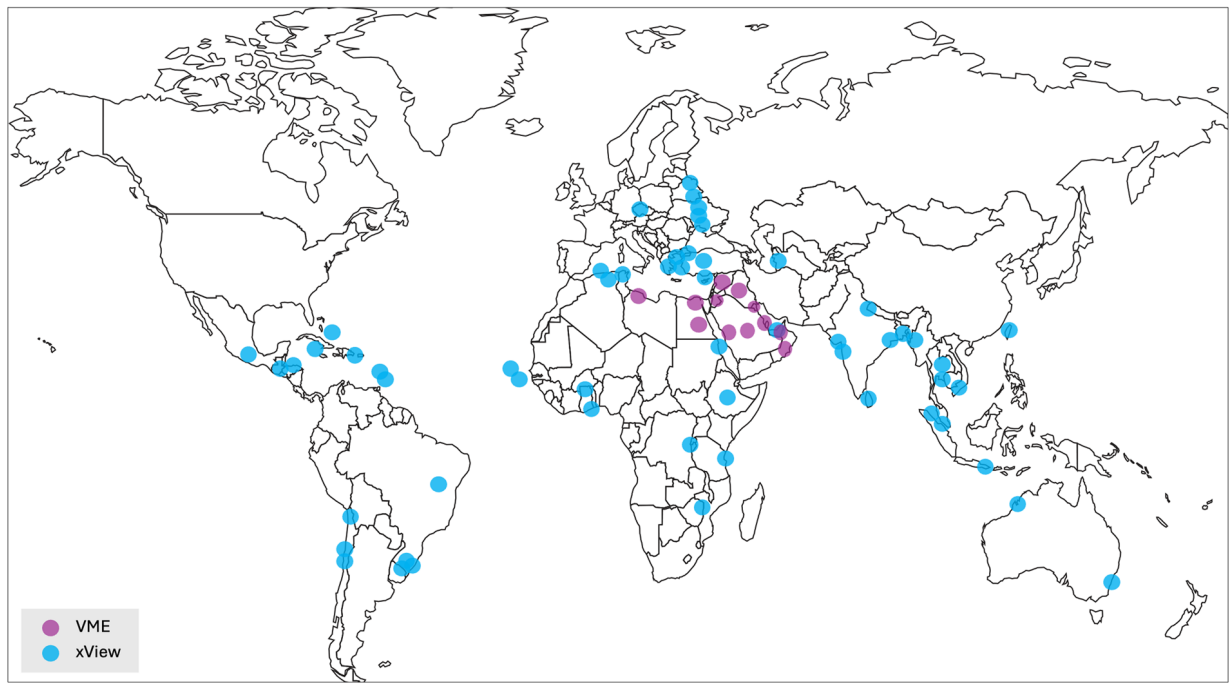

**Fig. 2** The geographical distribution of VME, and xView, denoted as purple and blue circles, respectively. There is no geographical information reported for the remaining datasets.

*Image Annotation.* After inspecting the taxonomies of the existing satellite imagery datasets for vehicle-related classes, we defined a three-class taxonomy comprising *car*, *bus*, and *truck* classes in our dataset. We decided to collect oriented bounding box (OBB) annotations as certain applications, such as traffic management, can leverage the direction information as well. We employed Co-one (https://www.co-one.co/), an AI and crowdsourcing-based data platform that promises 95% annotation accuracy. The annotation process started with preparing the guideline handbook, which outlined the project overview, technical guidelines, targeted categories with definitions and examples, rules and tips for the annotation process, and the deliverable format. Then, the data annotation process was conducted with a crowdsource of 6000+ people, where each group focused on a specific category. Finally, an annotation review process was implemented to detect mislabeled annotations via a cross-validation system; and an expert was employed to correct such cases. After the annotation quality review process, final annotations were delivered. We provided images in lossless PNG format, and received OBB annotations in YOLO format as text (*.txt) files. Each annotation file contains the image name, and each line in the file represents a targeted object as follows: $x_1, y_1, x_2, y_2, x_3, y_3, x_4, y_4$, *category_id*, where $(x_1, y_1)$ is the top left, $(x_2, y_2)$ is the top right, $(x_3, y_3)$ is the bottom right, and $(x_4, y_4)$ is the bottom left point of OBB, and *category_id* indicates the class index as 0, 1, 2 corresponding to *car*, *bus*, and *truck*, respectively. Additionally, we obtained standard horizontal bounding box (HBB) annotations based on the minimum and maximum $x$ and $y$ coordinates of the OBB annotations with their category. To better help the community utilize the dataset, we provide both the oriented and horizontal bounding box annotation files.

*Final Dataset.* Out of 4,303 images annotated, 21 images were deemed damaged or corrupted and excluded from the dataset. Hence, the final dataset contains 4,282 images with a total of 113,737 objects comprising 101,564 cars, 5,327 buses, and 6,846 trucks while 241 images do not contain any instances of the target object classes and are tagged as *no_label*. The distribution of classes is shown in Fig. 3a. Also, Fig. 3b,c,d highlight the area distribution of cars, buses, and trucks in pixels, respectively. It is observed that all of the car instances fall under the small object range (i.e., area (pixels) $< 32^2$) defined in the MS-COCO evaluation, specifically within the first half of the range (i.e., area (pixels) $< 512$) which is considered tiny objects. On the other hand, both the bus and truck instances fall mostly within the small object range (area (pixels) $< 32^2$), with almost negligible overlap into the medium object range ($32^2 <$ area (pixels) $< 96^2$). We provide training, validation, and test sets of the dataset following a random split with a ratio of 5/8, 1/8, and 2/8, respectively. Table 1 presents the statistics for all VME categories, outlining the details of each split.

**CDSI Dataset.** This section introduces the related object detection datasets in satellite imagery, such as xView, DOTA-v2.0, VEDAI, FAIR1M-2.0, and DIOR. Also, it describes the filtering and consolidation process of the CDSI dataset.

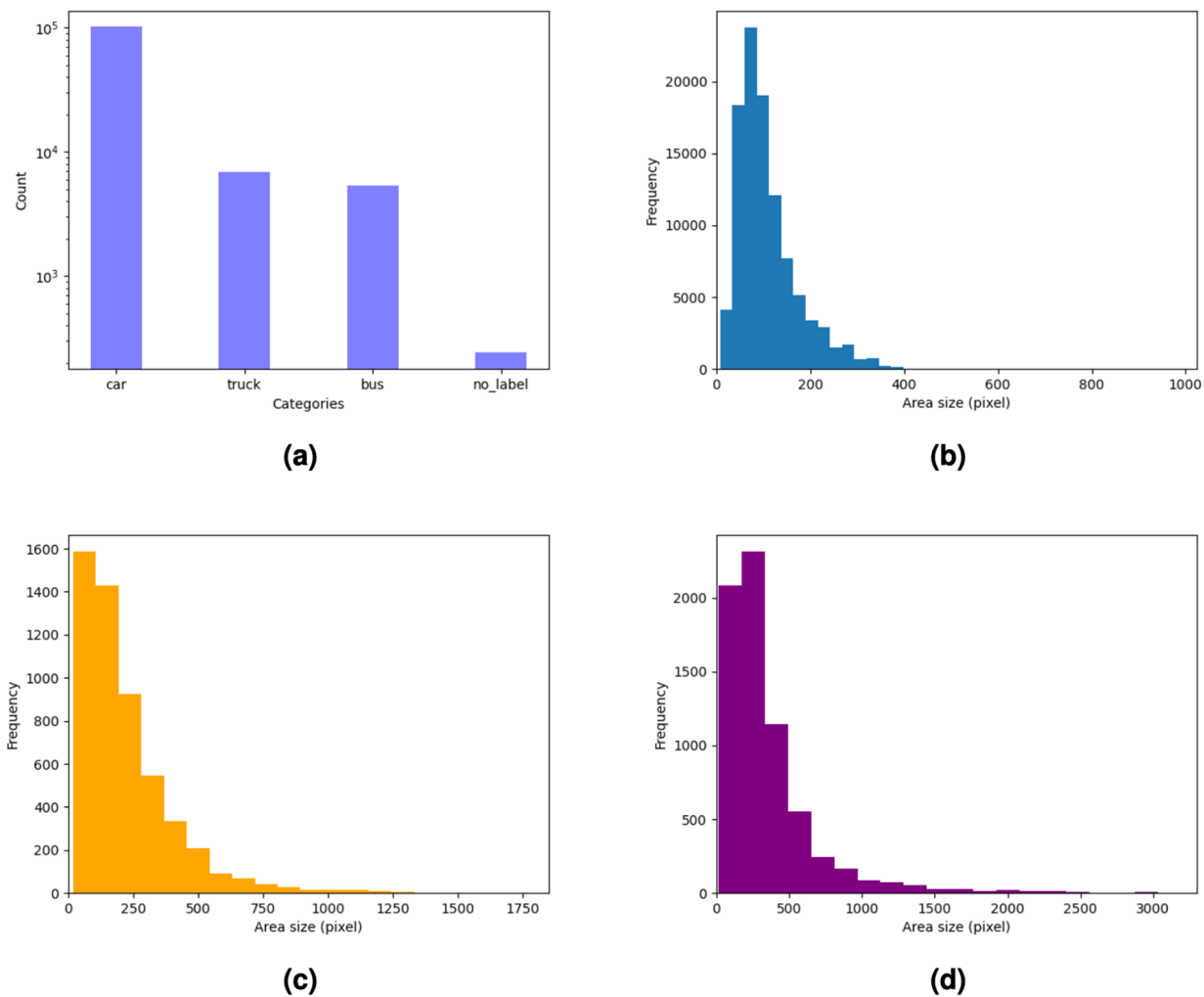

**Fig. 3** Statistical properties of the object categories in the VME dataset. (**a**) Distribution of VME categories, (**b**) Area distribution of cars, (**c**) Area distribution of buses, (**d**) Area distribution of trucks.

|  | Training | | Validation | | Test | |
| --- | --- | --- | --- | --- | --- | --- |
| **Categories** | # ann. | # imgs | # ann. | # imgs | # ann. | # imgs |
| Car | 63,051 | 2,449 | 12,055 | 510 | 26,458 | 988 |
| Bus | 3,079 | 964 | 674 | 183 | 1,574 | 394 |
| Truck | 4,140 | 1,041 | 1,004 | 225 | 1,702 | 428 |
| All | 70,270 | 2,505 | 13,733 | 525 | 29,734 | 1,011 |

**Table 1.** Number of images and annotations in each category across training, validation, and test splits of the VME dataset.

*Existing Datasets.* We explored a large list of publicly available datasets for object detection to employ in our study. We excluded those with low-altitude, drone-based, and UAV-based datasets and the datasets with high ground sample distance (GSD) ranges or hidden contexts, such as COWC[33], PaCaBa[34], PSU[35], and VisDrone[36]. Even some of the new datasets are not yet released, e.g., VehSat[37] and EAGLE[38]. The following are the datasets we employed in our study.

**xView**[22] is considered one of the largest publicly available datasets containing 846 images collected from Maxar at various locations around the world, as shown in Fig. 2. The images are available with 30cm/pixel spatial resolution, and the average dimension of the images is 3316 × 2911. The dataset has 60 object classes with 1 million object instances annotated using horizontal bounding boxes for all splits, while the ground truth of testing split is not available. The xView repository (https://challenge.xviewdataset.org/data-download) provides the training and validation images in TIF format and the annotations in GeoJSON format.

**DOTA-v2.0**[20] contains 2,423 images gathered from Google Earth, different satellites supplied by the Resources Satellite Data and Application Center in China, and aerial images supplied by CycloMedia B.V. The size of the images ranges from 800 to 20,000 pixels, and their spatial resolution varies between 0.1m/pixel to 4.5m/pixel. The dataset contains 18 object classes and objects are annotated using both oriented and horizontal bounding boxes. The dataset is presented in three versions, where the final version (v2.0) contains a total of 1,793,658 object instances of all splits, while the ground truth of the testing split is not available. The DOTA images were released with no geographical information. We obtained DOTA-v2.0 from its repository (https://captain-whu.github.io/DOTA/index.html); the images are in PNG format and its annotations are in YOLO (TXT) format. Acquiring DOTA-v2.0 requires the users to download DOTA-v1.0 first, then get the v2.0.

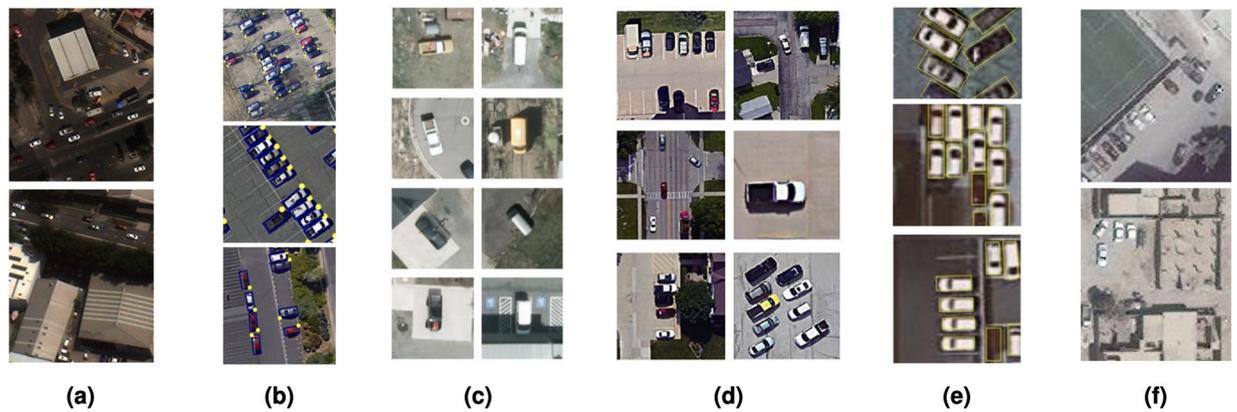

**Fig. 4** Example images with car-related objects in (**a**) xView[22], (**b**) DOTA-v2.0[20], (**c**) VEDAI[21], (**d**) DIOR[29], (**e**) FAIR1M-2.0[30], (**f**) VME (our) datasets.

**VEDAI**[21] was built specifically for detecting vehicles in satellite imagery, such as boats, planes, tractors, cars, and vans. The dataset provides two sets of 1,246 images in colored and infrared format, each set at a different spatial resolution (12.5cm/pixel or 25cm/pixel), and hence, image dimensions (1024 × 1024 or 512 × 512 pixels). The annotation format used for the dataset is the oriented bounding box. No geographic information is revealed in VEDAI. In our study, we downloaded colored images with a spatial resolution of 25cm/pixel (i.e., image dimensions of 512 × 512 pixels) from the VEDAI repository (https://downloads.greyc.fr/vedai/). The annotations are stored in TXT files, reporting the four corners of OBBs with the category.

**DIOR**[29] is another large-scale benchmark dataset for object detection in optical satellite images. It consists of 23,463 images annotated for 20 object categories and 192,512 object instances using horizontal bounding boxes. Spatial resolution of images is between 0.5m/pixel and 30m/pixel. The dataset claims to cover more than 80 countries, but the specific list of countries has not been released. The dataset can be downloaded from DIOR repository (https://gcheng-nwpu.github.io/); which delivers the images in JPG format and the annotation files in PASCAL-VOC (XML) format.

**FAIR1M-2.0**[30] contains more than 20,000 images with more than 1 million instances of fine-grained object categories. The images are gathered from Google Earth and the Gaofen satellites, with spatial resolutions between 0.3m/pixel and 0.8m/pixel. The object annotations were collected for five main categories and 37 sub-categories using oriented bounding boxes. It is stated that the dataset covers different continents; but the country- or city-level details about the image locations are not published. The dataset can be obtained from FAIR1M repository (https://gaofen-challenge.com/benchmark); its annotation files are presented in PASCAL-VOC (XML) format, and the images are offered in TIF format.

*Category Mapping.* To construct a unified benchmark dataset for car detection in satellite imagery, we investigated the taxonomies of the aforementioned datasets. Each dataset labels car-related objects differently, using terms like "small car," "small vehicle," "vehicle," "car," or "van." Thus, we visually inspected these categories to ensure they correspond to the same "car" object we are targeting. For instance, "small car" in xView and "small vehicle" in DOTA-v2.0 refer to standard cars, while in DIOR "vehicle" covers a broader range of vehicles (e.g., cars, trucks, buses, and vans), with "car" being a subset of this general category. Figure 4 illustrates the car-related objects across datasets that we target for constructing the CDSI dataset. We conclude that these classes can be mapped to the same object type, i.e., car, with certain conditions such as filtering by typical car size. Specifically, car-related categories were mapped to the *car* category in CDSI for objects with an HBB area of less than 400 pixels, as detailed in Table 2. To avoid the challenges associated with training an object detection model using only a single class, we ensured the model encountered hard negatives–objects similar in size to cars. To achieve this, we opted to group all other small objects into a single category called "other." Similarly, instances from all other categories with an HBB area of less than 400 pixels were mapped to the *other* small object category in CDSI, as reported in Table 2. As a result, the CDSI dataset consists of two classes: "car" and "other." The details about data processing and filtering steps are explained next.

*Data Processing and Filtering.* Each dataset uses a different annotation style (e.g., OBB, HBB, or both) and adopts different data representation and file format (e.g., XML files with PASCAL-VOC format, TXT files with YOLO format, JSON files with MS-COCO format, etc.). To consolidate all of the datasets, we designed a data processing pipeline, illustrated in Fig. 5, with the following steps:

| Dataset | Annotation | Image Resolution (m/pixel) | Original Categories | All Objects | All Images | Car-related Classes | Car-related Objects | Other Small Objects | Retained Images |
|---|---|---|---|---|---|---|---|---|---|
| xView | HBB | 0.3 | 60 | 601,718 | 846 | small car | 210,184 | 55,752 | 752 |
| DOTA-v2.0 | HBB/OBB | 0.1 to 4.5 | 18 | 349,675 | 2,423 | small vehicle | 175,160 | 37,037 | 1,300 |
| VEDAI | OBB | 0.25 | 10 | 3,754 | 1,246 | car, van | 1,422 | 1,292 | 1,057 |
| DIOR | HBB/OBB | 0.5 to 30 | 20 | 192,512 | 23,463 | vehicle | 23,964 | 33,521 | 5,327 |
| FAIR1M-2.0 | OBB | 0.3 to 0.8 | 37 | 594,482 | 24,775 | small car, van | 384,488 | 48,416 | 10,795 |
| CDSI* (our) | HBB | 0.1 to 30 | 2 | 971,236 | 19,321 | car | 795,218 | 176,018 | 19,321 |
| VME (our) | OBB/HBB | 0.3, 0.4, 0.5 | 3 | 113,737 | 4,041 | car | 101,542 | 9,601 | 4,019 |
| CDSI (our) | HBB | 0.1 to 30 | 2 | 1,082,379 | 23,250 | car | 896,760 | 185,619 | 23,250 |

**Table 2.** Statistics of car-related and other small object categories in different datasets. CDSI* indicates the version of CDSI without VME.

- **Annotation standardization**: We standardize all the annotations from different datasets to HBB style. Then, we convert standardized annotations into MS-COCO format which is defined by four values in pixels *(x_min, y_min, width, height)*.
- **Car-related object size filtering**: Given that we are interested in a GSD range of 30-50cm per pixel, we assume that an object with an area greater than 400 pixels is unlikely a car. To verify this assumption, we analyze the car size distributions in all datasets as in Fig. 6. This analysis reveals that an area size of less than 400 pixels reports for more than 90% of all car-related object instances across all datasets. Therefore, we decided to filter out all object instances with an area larger than 400 pixels; even if they were originally labeled as cars. During our visual inspection, we discovered that these cases often relate to labeling errors or images with spatial resolutions exceeding the targeted GSD range.
- **Relabeling small objects**: Using the same threshold, we repeat Step 2 to identify all other small object instances with an area less than 400 pixels and label them as "other" category.
- **Training setups**: Depending on the experimental setup, the car-related object instances are merged with the other small object instances to construct *car-other* setup for the model training. In contrast, only car-related objects are employed to form *car* setup for the model training (refer to the "Technical Validation" section for details).

*Final Dataset.* Table 2 provides general information and summary statistics about all datasets (individual or consolidated) before and after the data processing and filtering pipeline. The final combined dataset, i.e., CDSI, contains a total of 23,250 images with 896,760 car-related object instances and 185,619 other small object instances. Note that we also created a version of CDSI, denoted as CDSI*, where we excluded VME dataset from the consolidation process to highlight the contribution of VME dataset. With regards to the training, validation, and test sets, we first created random splits of all images in each individual dataset after filtering with a ratio of 5/8, 1/8, and 2/8 (as in the VME dataset). We then combined the resulting splits from different datasets to form the final data splits for both CDSI and CDSI* datasets. For instance, the CDSI training set is simply a union of training sets of all datasets, and the same rule applies for validation and test sets.

## Data Records
The repository available at Zenodo[39] consists of (a) the VME dataset, including satellite images and annotation files, and (b) the scripts and instructions for creating the CDSI dataset.

**Overview of the repository files and their formats.** The repository is structured into four components as follows:

- *annotations_OBB*: This folder holds TXT files in YOLO format with Oriented Bounding Box (OBB) annotations. Each annotation file is named after the corresponding image name, with each line describing a targeted object as follows: $x_1, y_1, x_2, y_2, x_3, y_3, x_4, y_4$, *category_id*, where $(x_1, y_1)$ is the top left, $(x_2, y_2)$ is the top right, $(x_3, y_3)$ is the bottom right, and $(x_4, y_4)$ is the bottom left point of OBB, and *category_id* indicates the class index as 0, 1, 2 corresponding to *car*, *bus*, and *truck*, respectively. The annotation files of images that do not include any of the targeted objects are empty.
- *annotations_HBB*: This folder contains HBB annotations in separate JSON files for training, validation, and test splits, formatted according to the MS-COCO standard defined by four values in pixels *(x_min, y_min, width, height)*.
- *satellite_images*: This folder contains VME images in PNG format, each with a resolution of 512 × 512 pixels.
- *CDSI_construction_scripts*: This directory contains all the necessary instructions for constructing the CDSI dataset, including: (a) guidelines for downloading each dataset from its respective repository, (b) scripts for converting each dataset to the MS-COCO format, located within the corresponding dataset folders, and (c) instructions for combining the datasets. The training, validation, and test splits are provided in the *CDSI_construction_scripts/data_utils* folder. Each split file lists the images from each dataset used in the car detection experiments for both detectors.

Additional information on the environment setup and required packages is available in the *README.md* file.

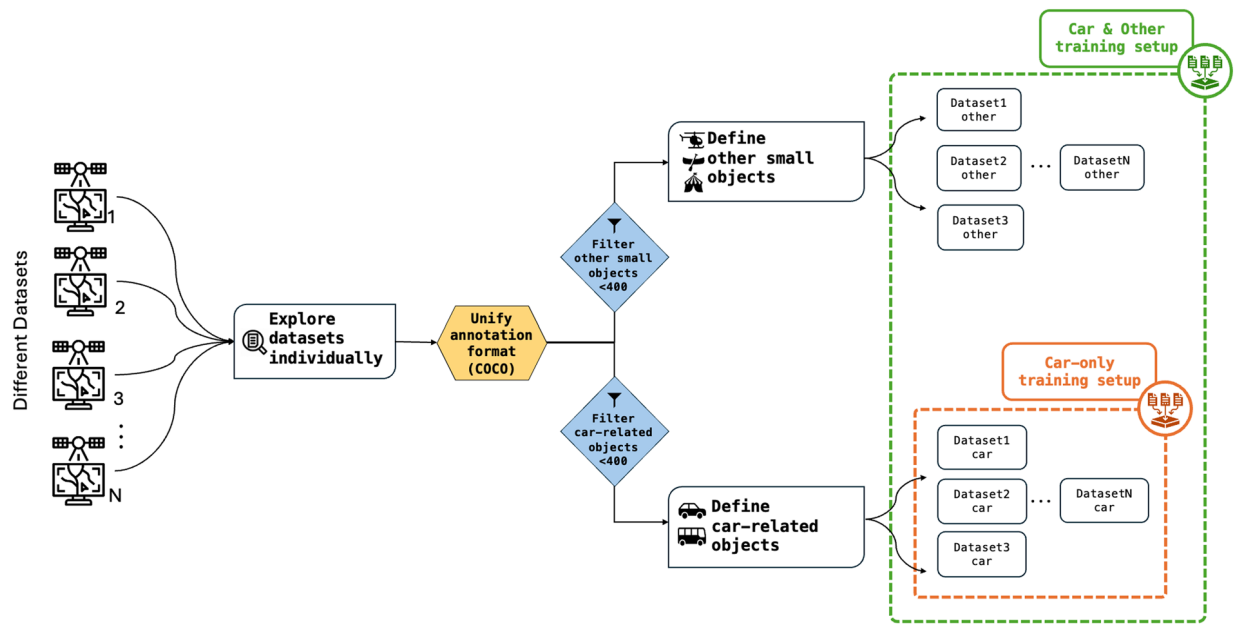

**Fig. 5** Dataset consolidation pipeline and final experimental setups.

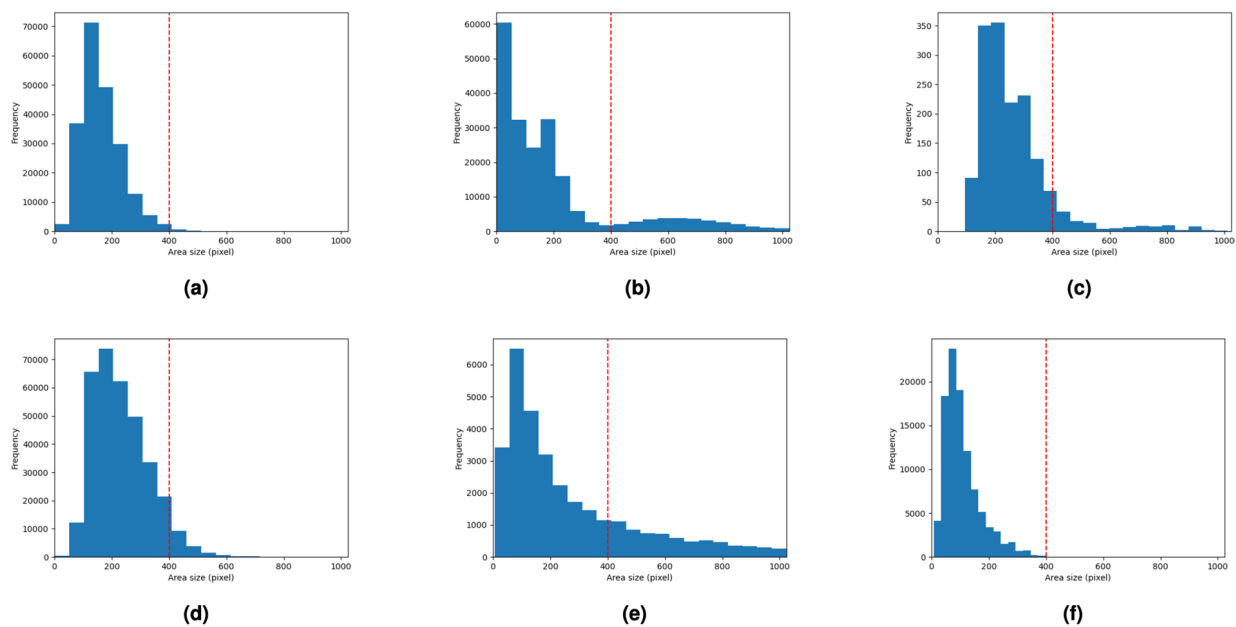

**Fig. 6** Distribution of car sizes in (**a**) xView, (**b**) DOTA-v2.0, (**c**) VEDAI, (**d**) FAIR1M-2.0, (**e**) DIOR, and (**f**) VME (our) datasets.

## Technical Validation

In this section, we perform a formal assessment of the quality of the VME annotations. Additionally, we provide details on benchmarks conducted across diverse setups and present analytical results to demonstrate the reliability and validity of the VME and CDSI datasets.

**VME Annotation Quality.** We implemented quality control to ensure the accuracy and consistency of the VME dataset annotations. We randomly selected around 5% of images across all 54 cities and resolutions and labeled these images in-house (by the lead author) to establish ground truth. This process yielded 5,664 ground truth annotations in 215 images. Next, we compared these labels with the annotations from the crowdsource-based platform to calculate True Positives (TP), False Positives (FP), and False Negatives (FN). We identified 5,496 TP, 7 FP, and 168 FN annotations. We then used these values to compute precision, recall, and F1 scores as 0.999, 0.970, and 0.984, respectively. Although some objects were missed, the crucial factor is that the identified objects are indeed the targeted ones, making the minimization of False Positives a priority. This process demonstrates that the annotations are highly accurate.

| Dataset | Training | | | | | | | Validation | | | | | | Test | |
|---|---|---|---|---|---|---|---|---|---|---|---|---|---|---|---|
| | original | | | car-other | | car | | original | | car-other | | car | | car | |
| | # cat. | # imgs | # ann. | # imgs | # ann. | # imgs | # ann. | # imgs | # ann. | # imgs | # ann. | # imgs | # ann. | # imgs | # ann. |
| xView | 60 | 524 | 380923 | 468 | 170280 | 432 | 135398 | 110 | 94786 | 101 | 46178 | 91 | 38146 | 167 | 36640 |
| DOTA-v2.0 | 18 | 1513 | 212720 | 796 | 123409 | 496 | 100250 | 317 | 55907 | 189 | 37907 | 113 | 33009 | 193 | 41901 |
| VEDAI | 10 | 772 | 2276 | 655 | 1650 | 451 | 900 | 162 | 477 | 137 | 367 | 95 | 178 | 183 | 344 |
| DIOR | 20 | 14547 | 119216 | 3302 | 34935 | 1958 | 15461 | 3050 | 23002 | 693 | 7034 | 411 | 2902 | 790 | 5601 |
| FAIR1M-2.0 | 37 | 14756 | 311558 | 6180 | 226757 | 5474 | 202187 | 1732 | 81613 | 1031 | 63513 | 870 | 57013 | 2969 | 125288 |
| CDSI* (our) | 2 | — | — | 11401 | 557031 | 8811 | 454196 | — | — | 2151 | 154999 | 1580 | 131248 | 4302 | 209774 |
| VME (our) | 3 | 2505 | 70270 | 2495 | 68721 | 2449 | 63038 | 525 | 13733 | 521 | 13364 | 509 | 12051 | 988 | 26453 |
| CDSI (our) | 2 | — | — | 13896 | 625752 | 11260 | 517234 | — | — | 2672 | 168363 | 2089 | 143299 | 5290 | 236227 |

Table 3. Statistics of the training, validation, and test splits in each experimental setup across datasets.

**Detection Benchmarks.** This section describes the benchmark setup and the application of state-of-the-art detection models to evaluate the technical quality and scientific significance of the VME and CDSI datasets. To this end, we explored three different setups to assess how varying number of images and objects (not necessarily cars) in a dataset affects the detection performance. In the first setup, we use the original datasets with their full taxonomy (i.e., all categories) to train object detection models. In the second setup, we use datasets containing only the images with instances of *car* and *other* small object categories. And, in the last setup, we use datasets containing only the images with *car* instances. To facilitate model training in the first setup, we created distinct training, validation, and test splits based on all the images in the *original* datasets using a ratio of 5/8, 1/8, and 2/8, respectively. In the second and third setups, these initial data splits were reduced to subsets containing only those images with relevant object instances. It is important to note that, at training time, we utilized each dataset's training and validation sets. However, at test time, we evaluated all trained models on the *car-only* test sets to obtain comparable performance scores. Table 3 presents the number of images and annotations across data splits and training setups (i.e., *original*, *car-other*, and *car*) for different datasets.

We conducted experiments using a state-of-the-art framework, Slicing Aided Hyper Inference (SAHI)[40]. SAHI is developed particularly for small object detection and provides a generalized slicing-aided inference and fine-tuning channel for detecting small objects. In SAHI, various object detectors were examined such as Fully Convolutional One-Stage Object Detection (FCOS)[41], VarifocalNet (VFNET)[42] and Task-aligned One-stage Object Detection (TOOD)[31]. In our study, we adopted the best-performing inference setup reported in SAHI, which is Slicing Aided Fine-tuning, Full-Inference, and Patch Overlap (SAHI+FI+PO) setting with TOOD detector from MMDetection library[43]. Additionally, we performed experiments with a more recent object detector called DINO[32] with Swin-L option from the MMDetection library following the SAHI+FI+PO inference setting. We trained a total of 22 models using the TOOD detector with a batch size of 16 for 24 epochs with SGD optimizer. For the *original* setup, we started training with a learning rate of 0.01 whereas, for the other setups, we started with a learning rate of 0.005. In all training setups, the learning rate was configured to change at epochs 9, 16, 22 with a learning rate decay equal to 0.1. Similarly, we trained a total of 22 models using the DINO Swin-L detector with a batch size of 2 for 36 epochs with AdamW optimizer with an initial learning rate of 0.0001, which was configured to change at epochs 27 and 33 with learning rate decay equal to 0.1. We ran all of our experiments on an NVIDIA A100 80GB GPU.

*VME Benchmark.* As we introduce our novel dataset for the first time, we perform experiments to provide baseline results. For this purpose, we train and test models with the original VME categories, utilizing both TOOD and DINO Swin-L detectors. Table 4 presents the class-specific and overall results obtained on the original VME test set. TOOD achieved an overall mAP50 score of 58.5% whereas DINO Swin-L achieved 62.7%. Notably, DINO Swin-L outperforms TOOD by 7.2%, with improvements of 6.2%, 5.9%, and 10.2% in mAP50 scores of car, bus, and truck categories, respectively. These baseline results highlight the challenging nature of the vehicle detection task and verifies our dataset's reliability for this challenging task. Given these results, we believe our novel dataset focused on Middle Eastern cities will play a key role in advancing vehicle detection in similar regions.

Figure 7 illustrates some examples of detection results from the baseline model applied to images from the Middle East sampled from the VME dataset. FP and FN are highlighted with yellow and magenta circles, respectively. To prevent clutter, detections for each object category are visualized separately. The results demonstrate the model's high detection accuracy, with occasional FP detections and rare FN occurrences, reflecting strong recall performance. These findings underscore the model's robustness while identifying opportunities for reducing FP rates.

*CDSI Benchmark.* This section provides a comprehensive benchmark across various datasets and setups, emphasizing the enhanced value introduced by the CDSI dataset. Additional analyses, including error evaluation and data visualization, further illustrate the strengths and limitations of the CDSI benchmark.

| Setup | TOOD | | | | | | | | DINO | | | | | | | |
|---|---|---|---|---|---|---|---|---|---|---|---|---|---|---|---|---|
| | Car | | Bus | | Truck | | All | | Car | | Bus | | Truck | | All | |
| | mAP | mAP50 | mAP | mAP50 | mAP | mAP50 | mAP | mAP50 | mAP | mAP50 | mAP | mAP50 | mAP | mAP50 | mAP | mAP50 |
| all cat. | 42 | 80.8 | 30.3 | 48.9 | 27.9 | 45.9 | 33.4 | 58.5 | 45.1 | 85.8 | 32.8 | 51.8 | 29.8 | 50.6 | 35.9 | 62.7 |

**Table 4.** VME baseline results obtained by training and testing the object detection models on the original VME data splits with all object categories as presented in Table 1. All mAP results are presented in percentage (%).

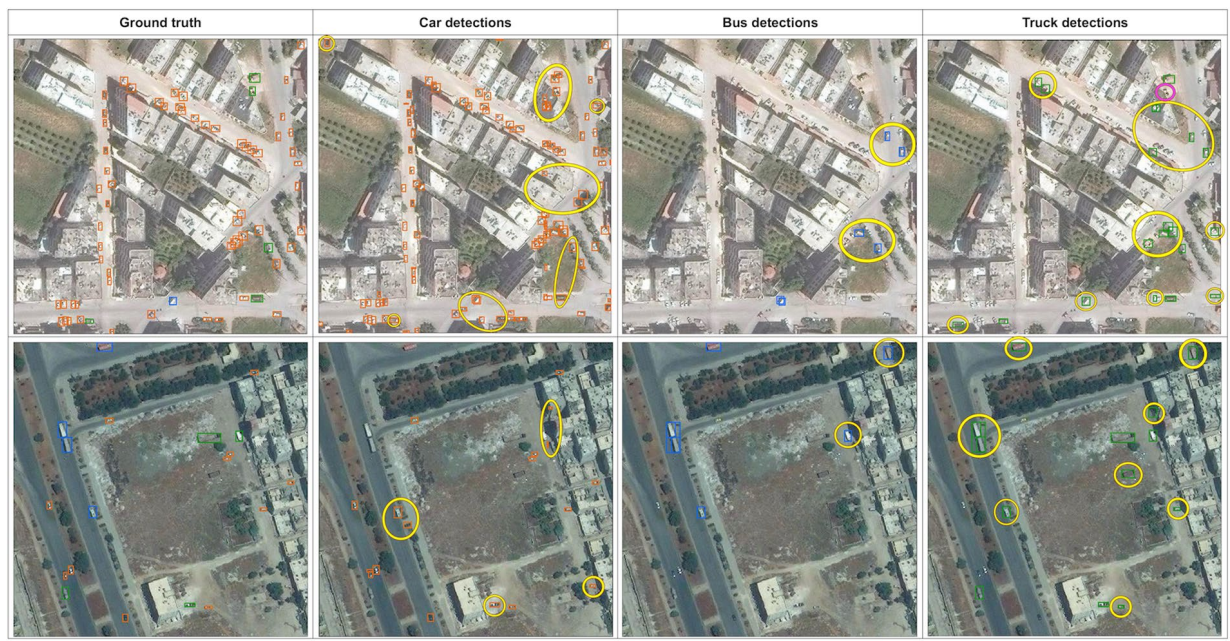

**Fig. 7** Detections on VME images employing VME baseline model trained on *all categories*. Yellow and magenta circles indicate examples of false positives and false negatives, respectively.

Table 5 summarizes the results achieved by both detectors, TOOD and DINO Swin-L, on CDSI and its constituents. Each row corresponds to a model trained on a particular dataset with a specific setup, e.g., *all categories*, *car-other*, and *car*. We evaluate each trained model on its own test set to quantify its *in-domain* performance as well as on the VME and CDSI test sets to assess its *generalization* capabilities. As highlighted before, we use *car-only* test sets in all cases for comparable results, which we discuss next. First, we observe that all the models trained on individual datasets exhibit poor performance on the VME dataset. Furthermore, the car detection performance does not improve even after combining all the existing datasets together (i.e., CDSI*). In essence, the models trained on existing datasets cannot effectively detect cars in images from the Middle East. In Fig. 8, the predictions of the VME *car* setup model are compared with the predictions of the models trained on xView and DOTA-v2.0 *car* setup on example images from the Middle East sampled from the VME dataset. The comparison shows that the models trained on xView and DOTA-v2.0 *car* setup struggle with detecting cars properly sometimes even in easy scenarios like cars on paved roads (top row).

Upon examining the CDSI dataset, the table presents the evaluation of predicting the CDSI test set across the trained models on each dataset individually, as well as the trained model on the CDSI. The results highlight the significance of the trained model on images from diverse sources, particularly in the context of detecting cars in satellite imagery. Additionally, the findings underscore the impact of incorporating the VME dataset in the trained model on *car* setup, revealing that the exclusion of VME in the trained model (CDSI*) leads to a decrease in mAP50 by 6% and 4.3% in TOOD and DINO Swin-L, respectively, when predicting on the CDSI test set.

To gain a deeper understanding of the significance of combining datasets (CDSI), we employed the Prithvi Foundation Model, a collaboration between IBM and NASA[44,45], which was pretrained on large-scale remote sensing data, including Harmonised Landsat Sentinel 2 (HLS). We utilized the IBM-NASA-Geospatial pretrained model with t-SNE (t-Distributed Stochastic Neighbor Embedding)[46], an unsupervised non-linear technique for visualizing feature embeddings, to explore how satellite images are represented in low-dimensional space based on their high-dimensional data. The t-SNE visualization helps in understanding the similarity between points, in this case, different satellite images from various datasets. The results, shown in Fig. 9, illustrate that the features of the FAIR1M-2.0 dataset are distinctly separate from the others. Additionally, xView shares some features with DOTA-v2.0 and DIOR, while VME shares certain features with DIOR and VEDAI. This outcome emphasizes the extent of training a model on a combined dataset for car detection (CDSI) and its implications for the field of car detection.

| Dataset | Setup | TOOD | | | | | | DINO | | | | | |
|---|---|---|---|---|---|---|---|---|---|---|---|---|---|
| | | Own Test Set | | VME Test Set | | CDSI Test Set | | Own Test Set | | VME Test Set | | CDSI Test Set | |
| | | mAP | mAP50 | mAP | mAP50 | mAP | mAP50 | mAP | mAP50 | mAP | mAP50 | mAP | mAP50 |
| xView | all cat. | 19.8 | 48.9 | 13.9 | 32.5 | 24.6 | 52.8 | 21.8 | 53.7 | 23.9 | 56.4 | 27.4 | 61.4 |
| | car-other | 19.6 | 49.3 | 16.3 | 39.6 | 21.7 | 51.5 | 19.8 | 50.1 | 22.4 | 53.4 | 27.2 | 60.5 |
| | car | 20.2 | 49.8 | 15.8 | 38.2 | 24.5 | 55.2 | 20.3 | 51.4 | 23.9 | 56.4 | 27.4 | 61.4 |
| DOTA-v2.0 | all cat. | 15.9 | 32.0 | 18.4 | 42.6 | 27.4 | 58.3 | 16.6 | 33.0 | 22.9 | 50.8 | 28.3 | 60.1 |
| | car-other | 18.1 | 35.8 | 17.6 | 44.4 | 26.6 | 58.2 | 17.7 | 36.0 | 17.9 | 39.5 | 28.5 | 60.6 |
| | car | 18.2 | 36.3 | 21.2 | 53.3 | 27.3 | 59.3 | 19.0 | 37.1 | 23.9 | 55.3 | 28.2 | 62.2 |
| VEDAI | all cat. | 63.5 | 89.7 | 2.3 | 4.8 | 14.5 | 37.6 | 55.1 | 89.1 | 1.7 | 4.2 | 15.0 | 40.1 |
| | car-other | 61.7 | 92.0 | 9.0 | 4.0 | 13.1 | 33.9 | 56.8 | 91.0 | 1.4 | 3.1 | 16.9 | 41.9 |
| | car | 47.1 | 81.7 | 2.4 | 5.1 | 15.7 | 40.2 | 52.3 | 87.7 | 1.9 | 4.7 | 15.4 | 40.3 |
| DIOR | all cat. | 29.2 | 59.4 | 12.1 | 32.5 | 20.2 | 47.9 | 28.9 | 58.1 | 17.2 | 41.7 | 23.8 | 53.3 |
| | car-other | 36.0 | 71.5 | 13.6 | 35.7 | 23.7 | 55.1 | 34.4 | 69.3 | 17.3 | 40.7 | 24.8 | 58.4 |
| | car | 30.9 | 65.9 | 12.0 | 33.5 | 20.9 | 51.9 | 31.5 | 65.8 | 20.2 | 47.7 | 26.7 | 60.1 |
| FAIR1M-2.0 | all cat. | 48.1 | 83.1 | 3.6 | 6.4 | 28.6 | 50.9 | 50.8 | 85.2 | 4.2 | 7.3 | 30.3 | 52.7 |
| | car-other | 52.0 | 90.3 | 5.5 | 10.9 | 32.5 | 59.3 | 52.0 | 90.7 | 5.5 | 9.7 | 32.2 | 59.0 |
| | car | 51.7 | 89.7 | 7.3 | 15.8 | 33.1 | 61.1 | 52.2 | 90.3 | 8.2 | 16.7 | 33.0 | 61.2 |
| CDSI* | car-other | 39.8 | 72.7 | 20.7 | 47.2 | 37.4 | 69.1 | 39.8 | 72.9 | 24.4 | 51.0 | 37.9 | 69.8 |
| | car | 39.9 | 72.5 | 22.0 | 50.0 | 37.9 | 69.6 | 39.9 | 72.8 | 29.3 | 62.3 | 38.7 | 71.3 |
| VME | all cat. | 39.8 | 76.3 | 39.8 | 76.3 | 25.5 | 53.0 | 44.5 | 84.0 | 44.5 | 84.0 | 29.6 | 60.7 |
| | car-other | 41.5 | 80.8 | 41.5 | 80.8 | 25.9 | 55.1 | 45.6 | 86.2 | 45.6 | 86.2 | 28.8 | 59.8 |
| | car | 42.2 | 81.2 | 42.2 | 81.2 | 25.9 | 54.5 | 45.8 | 86.5 | 45.8 | 86.5 | 27.9 | 58.8 |
| CDSI | car-other | 40.6 | 73.8 | 43.3 | 82.3 | 40.6 | 73.8 | 40.6 | 74.5 | 46.1 | 86.8 | 40.6 | 74.5 |
| | car | 40.5 | 73.8 | 43.0 | 81.9 | 40.5 | 73.8 | 40.7 | 74.4 | 45.7 | 86.4 | 40.7 | 74.4 |

**Table 5.** Experimental results achieved by TOOD and DINO Swin-L detectors trained on various datasets under different setups. The evaluation results are obtained on each dataset's own *car-only* test set, VME *car-only* test set, and CDSI *car-only* test set with SAHI+FI+PO inference setting. All mAP results are presented in percentage (%).

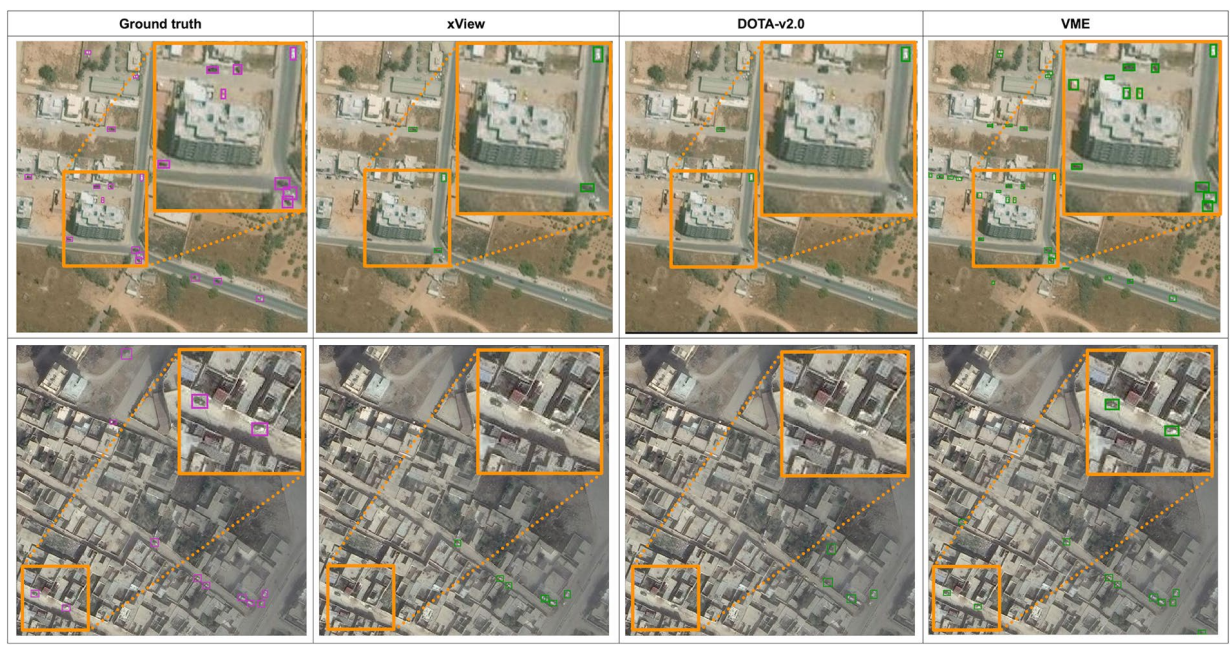

**Fig. 8** Comparison of detections on VME images employing the model trained on VME *car* setup versus detections of the models trained on xView and DOTA-v2.0 *car* setup.

To delve deeper into the details of Table 5, we analyze the performance across various setups using individually trained datasets, VME and CDSI, and assess how their predictions performed on their respective

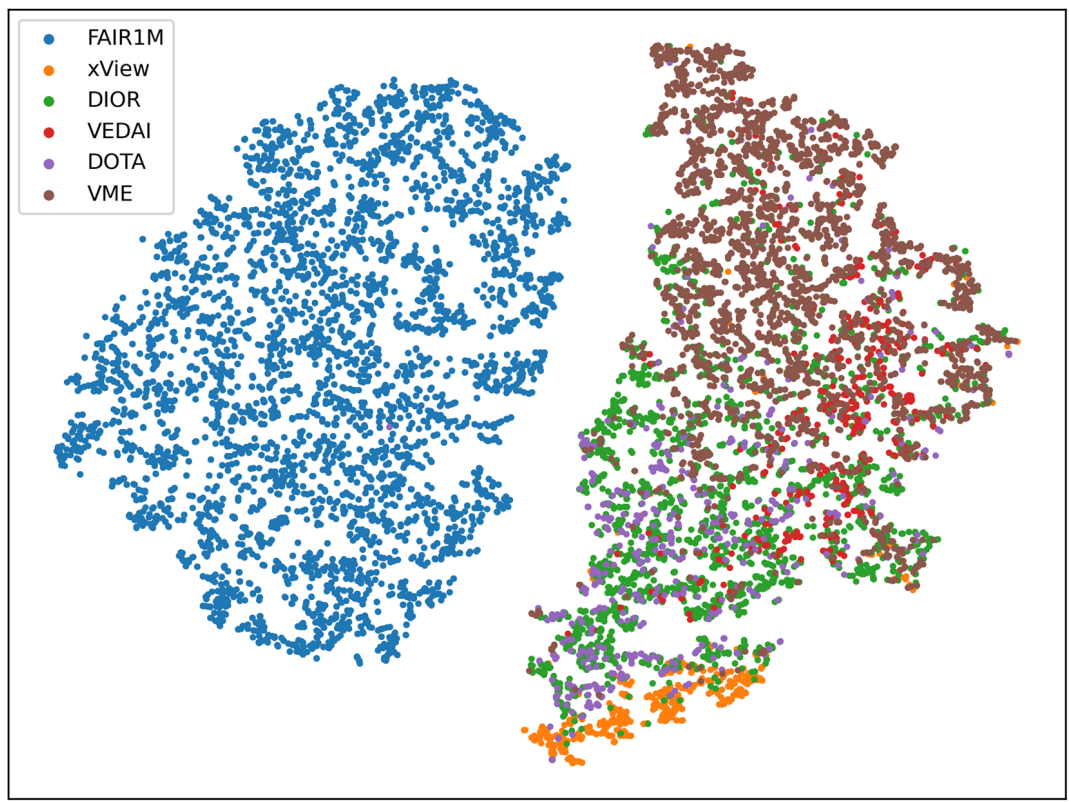

**Fig. 9** t-SNE visualization of the proposed CDSI dataset.

*car-only* test sets, VME *car-only* test set and CDSI test set. Overall, the *car* setup performed better with the TOOD detector in most setups, except for VEDAI and DIOR, which produced better results in the *car-other* setup in terms of mAP50(%). When focusing on TOOD models trained on other datasets in the *car* setup and evaluated on the VME test set, results reveal poor performance. Notably, low mAP50 scores were observed for models trained on VEDAI (5.1%) and FAIR1M-2.0 (15.8%), likely due to VEDAI's limited number of images and annotations, which struggled with car detection in images with varied resolutions and higher car densities. Despite FAIR1M-2.0 being the largest dataset in terms of car-related objects and images, Fig. 9 indicates that its image features differ significantly from those of the VME dataset. A similar pattern is seen in *all categories* and *car-other* setups for all models. On the other hand, DINO Swin-L shows slight improvement across all trained models, mirroring the pattern observed with TOOD. Notably, the model trained on CDSI in the *car-other* setup achieved the highest mAP50 score (86.8%) on the VME test set.

To investigate the root causes of errors, we perform an analysis of the detection results[47] from the DINO Swin-L model trained on VME and CDSI using the *car-other* setup. Figures 10 and 11 show a breakdown of errors for the *car* class for VME and CDSI, respectively. The error analysis provides various insights to identify areas for improvement, including: 1) IoU thresholds of 0.75, 2) IoU thresholds of 0.50, 3) post-localization error removal, 4) false positives within supercategories, 5) category confusion, 6) background false positives, and 7) false negatives, represented as C75, C50, Loc, Sim, Oth, BG, and FN, respectively. Note that the area under each Precision-Recall curve is shown in brackets in the legend. In the case of VME (Fig. 10), overall AP at IoU=.75 is 0.432 (C75), and simply lowering IoU=0.5 increases the AP to 0.861 (C50), whereas perfect localization could increase AP to 0.898 (Loc). We observe some error due to the confusion between the *car* and *other* categories and removing such class confusions would only raise AP slightly to 0.909 (Oth). However, we see a bigger room for improvement by eliminating background false positives (i.e., confusions with other small background objects), which boosts the AP to 0.99 (BG). Surprisingly, in the case of VME, the model does not suffer too much from false negatives (i.e., missed detections). On the other hand, for the model trained on CDSI (Fig. 11), we see similar trends in general regarding the errors due to category confusions and background false positives. However, resolving such issues can boost AP to a maximum of 0.851 (BG), which means the rest of the errors are missing detections. The missed detections in the model trained on CDSI are due to the diversity in object instances and variations in image characteristics collected from different regions. In summary, both plots illustrate that the errors are dominated by imperfect localization and background confusions.

## Usage Notes

The VME dataset and the script for creating the CDSI dataset are available at Zenodo[39]. VME images are available in resolutions ranging from 30 to 50 cm per pixel. However, the climate conditions in the Middle East, including haze and airborne dust, can affect the clarity of these images. As a result, some images may have a blurry appearance or exhibit reflections.

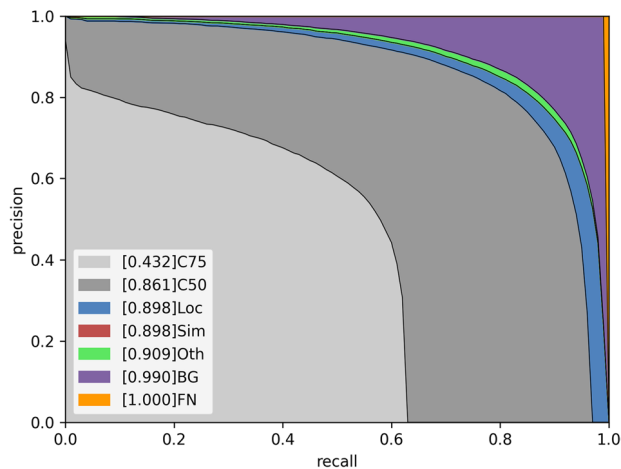

**Fig. 10** Error analysis for the *car* category of the DINO Swin-L detector trained on VME using the *car-other* setup.

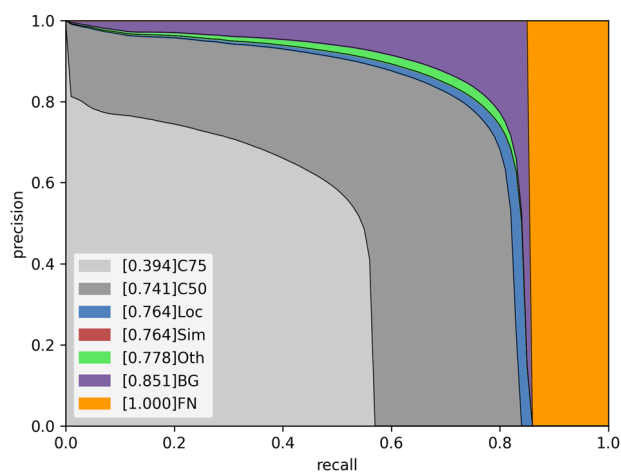

**Fig. 11** Error analysis for the *car* category of the DINO Swin-L detector trained on CDSI using the *car-other* setup.

### Code availability
The data preprocessing script for constructing CDSI dataset, which is written in Python, is available on Zenodo[39] and GitHub repository (https://github.com/nalemadi/VME_CDSI_dataset_benchmark) under *CDSI_construction_scripts* folder. The file *README.md* provides detailed instructions for building the CDSI dataset, which includes downloading the datasets, converting each to MS-COCO format, and explaining the combination mechanism. Each subfolder is named after its corresponding dataset and contains a conversion script to MS-COCO format. All the required Python packages are listed in the *requirements.txt* file located within the *CDSI_construction_scripts* folder.

### Acknowledgements
This publication was made possible by GSRA grant, I.D. # GSRA7-1-0421-20022, from the Qatar National Research Fund (a member of Qatar Foundation). We sincerely thank our colleague Masoomali Fatehkia (Qatar Computing Research Institute, HBKU) for assisting with image collection. Ingmar Weber is supported by funding from the Alexander von Humboldt Foundation and its founder, the Federal Ministry of Education and Research (Bundesministerium für Bildung und Forschung).

### Author contributions
N.A. conceived the dataset collection and preparation, dataset validation, and pre-processing, conducted the experiments, and wrote the manuscript. I.W. and F.O. facilitated access to Middle East satellite imagery. F.O. provided analysis techniques. I.W., Y.Y., and F.O. supervised and guided the study. All authors reviewed the manuscript.

### Competing interests
The authors declare no competing interests.

### Additional information
**Correspondence** and requests for materials should be addressed to N.A.-E.

**Reprints and permissions information** is available at www.nature.com/reprints.

**Publisher's note** Springer Nature remains neutral with regard to jurisdictional claims in published maps and institutional affiliations.

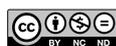 **Open Access** This article is licensed under a Creative Commons Attribution-NonCommercial-NoDerivatives 4.0 International License, which permits any non-commercial use, sharing, distribution and reproduction in any medium or format, as long as you give appropriate credit to the original author(s) and the source, provide a link to the Creative Commons licence, and indicate if you modified the licensed material. You do not have permission under this licence to share adapted material derived from this article or parts of it. The images or other third party material in this article are included in the article's Creative Commons licence, unless indicated otherwise in a credit line to the material. If material is not included in the article's Creative Commons licence and your intended use is not permitted by statutory regulation or exceeds the permitted use, you will need to obtain permission directly from the copyright holder. To view a copy of this licence, visit http://creativecommons.org/licenses/by-nc-nd/4.0/.